\documentclass[conference]{IEEEtran}
\IEEEoverridecommandlockouts
\usepackage{cite}
\usepackage{amsmath,amssymb,amsfonts}
\usepackage{algorithmic}
\usepackage{graphicx}
\usepackage{textcomp}
\usepackage{xcolor}

\def \iou{\mathcal{L}_{IoU}}
\def \imean{\overline{\iou}}
\def \r{\mathcal{R}}

\def \wiou#1{\mathcal{L}_{WIoUv{#1}}}
\def \out{\frac{\iou^*}{\ \imean\ }}

\def \cenconnect{(x-x_{gt})^2 + (y-y_{gt})^2}
\def \diagsqr{W^2_g + H^2_g}

\def\BibTeX{{\rm B\kern-.05em{\sc i\kern-.025em b}\kern-.08em
    T\kern-.1667em\lower.7ex\hbox{E}\kern-.125emX}}
\begin{document}

\title{Student Classroom Behavior Detection based on Improved YOLOv7\\
{\footnotesize \textsuperscript{*}Note: Sub-titles are not captured in Xplore and
should not be used}
\thanks{Identify applicable funding agency here. If none, delete this.}
}

\author{\IEEEauthorblockN{1\textsuperscript{st} Fan Yang}
\IEEEauthorblockA{\textit{Jinan University} \\
Guangzhou, China \\
winstonyf@qq.com}
\and
\IEEEauthorblockN{2\textsuperscript{nd} xiaofei wang}
\IEEEauthorblockA{\textit{School of films and animation of the college of Chinese \& ASEAN arts} \\
\textit{Chengdu University}\\
ChengDu, China \\
wangxiaofei@cdu.edu.cn}

}

\maketitle

\begin{abstract}
Accurately detecting student behavior in classroom videos can aid in analyzing their classroom performance and improving teaching effectiveness. However, the current accuracy rate in behavior detection is low.      To address this challenge, we propose the Student Classroom Behavior Detection method, based on improved YOLOv7. First, we created the Student Classroom Behavior dataset (SCB-Dataset), which includes 18.4k labels and 4.2k images, covering three behaviors: hand raising, reading, and writing. To improve detection accuracy in crowded scenes, we integrated the biformer attention module and Wise-IoU into the YOLOv7 network. Finally, experiments were conducted on the SCB-Dataset, and the model achieved an mAP@0.5 of 79$\%$, resulting in a 1.8$\%$ improvement over previous results. The SCB-Dataset and code are available for download at: https://github.com/Whiffe/SCB-dataset.
\end{abstract}


\section{Introduction}
Behavior detection technology\cite{b1} has made it possible to analyze student behavior in class videos,  it can provide information on the classroom status and learning performance of students, making it an essential tool for teachers, administrators, students, and parents in schools. However, in traditional teaching methods, it is difficult for teachers to observe the learning situation of each individual student, school administrators typically rely on on-site observations and student performance reports to identify educational problems, while parents depend on communication with teachers and students to understand their child's learning situation. Therefore, utilizing behavior detection technology to accurately detect and analyze student behavior can provide comprehensive and accurate feedback for education and teaching.

Existing student behavior detection algorithms can be classified into three categories: video-action-recognition-based\cite{b2}, pose-estimation-based\cite{b3}, and object detection-based\cite{b4}, with the latter being a promising solution due to recent breakthroughs.   Although video-based detection allows for the recognition of continuous behavior, it requires a large number of annotated samples such as in the AVA dataset\cite{Ava} for SlowFast\cite{b6} detection which includes 1.58 million annotations.   Video behavior recognition is still under development, and in some cases, actions can only be determined by context or scene alone as seen in UCF101\cite{b7} and Kinetics400\cite{b8}.   Pose-estimation algorithms obtain joint position and motion information but are not adequate for detecting behavior in overcrowded classrooms.   Based on the challenges at hand, an object-detection-based approach, such as YOLOv7\cite{b9}, has been employed to analyze student behavior in this paper.

Object detection networks have shown impressive results on public datasets\cite{b10,b11}. Nevertheless, classroom behavior detection has some challenges such as variations in behavior occurring in different environments, among different people, and from different angles, as well as the occlusion of students compared to popular object detection datasets like MS COCO \cite{coco}.

In this study, we investigate the potential of object detection to automatically identify important student behaviors in the classroom, such as hand-raising, reading, and writing. To this end, we constructed the Student Classroom Behavior Dataset (SCB-Dataset). The SCB-Dataset fills a gap in current research on detecting student behavior in classroom teaching scenes. We have conducted extensive data statistics and benchmark tests to ensure the quality of the dataset, providing reliable training data.

In this study, we investigate the potential of object detection networks to automatically identify important student behaviors in the classroom, such as hand-raising, reading, and writing.   To this end, we constructed the Student Classroom Behavior Dataset (SCB-Dataset).   The SCB-Dataset fills a gap in current research on detecting student behavior in classroom teaching scenes.   We have conducted extensive data statistics and benchmark tests to ensure the quality of the dataset, providing reliable training data.   Additionally, we did not include standing, sitting, and speaking behaviors in the dataset as they are already covered in the AVA dataset, which we can use directly.

\begin{figure*}
\centerline{\includegraphics[width=1\textwidth]{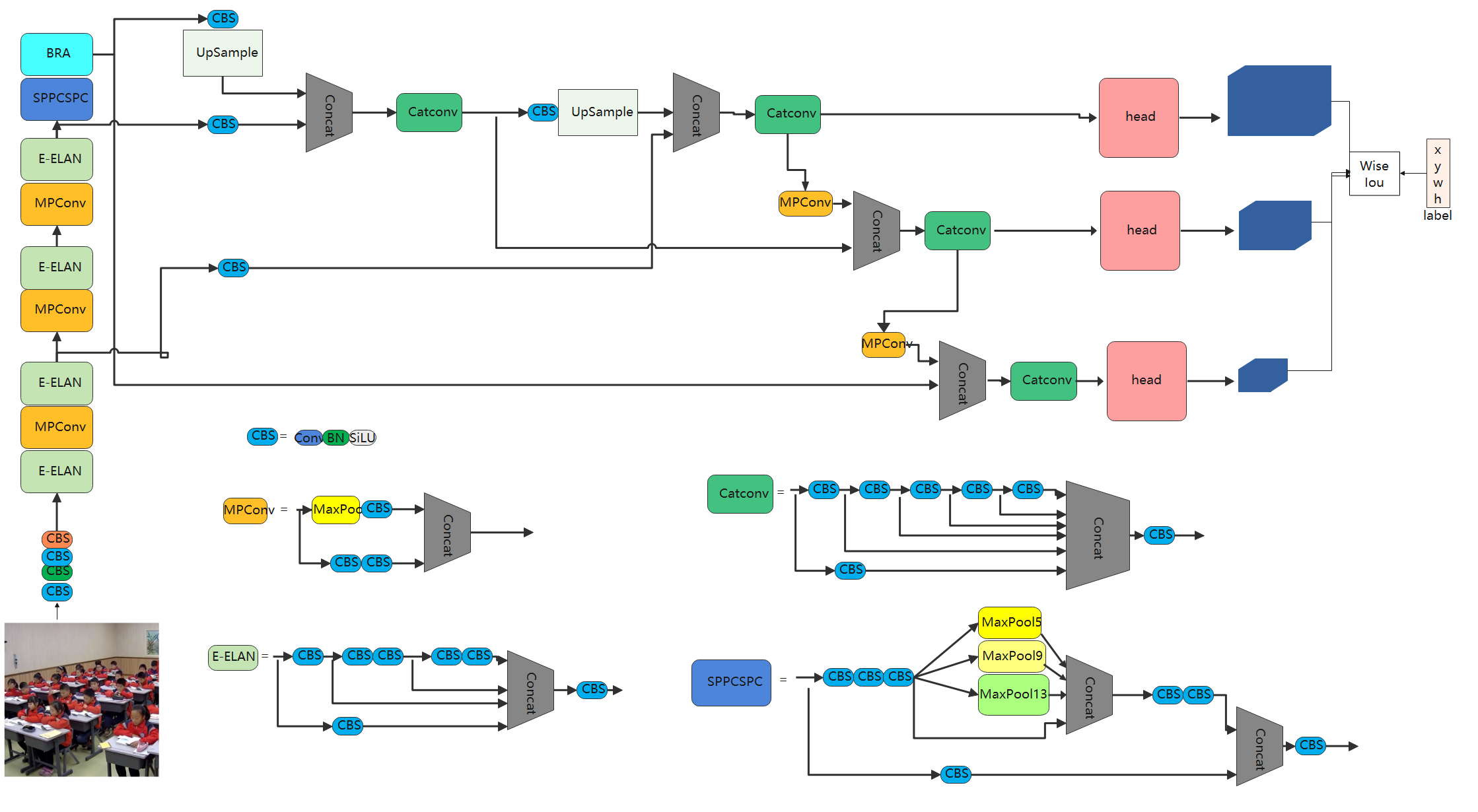}}
\caption{The framework of student classroom behavior detection based on improved YOLOv7.}
\label{YOLOv7+BRA+wise}
\end{figure*}

YOLOv7 is one of the best one-stage object detection algorithms currently available, and we attempted to train it on SCB-Dataset for better detection results. However, we found that the original version of YOLOv7 still had some room for improvement after training - for example, it would misidentify other actions as raising hands and fail to detect smaller hand-raising actions. Moreover, in areas of dense student's scenes or occlusion between students, the Bounding Box generated by the algorithm would often display some errors. To address these issues, we incorporated two novel methods , as shown in Fig.~\ref{YOLOv7+BRA+wise}. First, we incorporated a dynamic sparse attention module called Bi-Level Routing Attention \cite{BiFormer} (BRA). In addition, we incorporated a dynamic non-monotonic FM uses the outlier degree , known as Wise-IoU \cite{Wise-IoU}, which successfully improved detection performance.

Our main contributions are as follows:

(1) This paper constructed a dataset of students raising hands in classrooms, enabling better research into classroom behavior detection. Moreover, the SCB-dataset fills a gap in behavior datasets in the educational field.  

(2) This paper proposes an improved model, named YOLOv7+BRA. We added a Bi-Level Route Attention module to the model to give it dynamic query-aware sparsity. Experimental results show that our method successfully improves detection accuracy and reduces false detection rates.

(3) This paper proposes an improved model named YOLOv7+Wise IoU that integrates the Wise-IoU loss function to improve the model's performance on datasets with uneven data quality. By dynamically allocating gradient gain based on the quality of anchor boxes, Wise-IoU reduces the number of training iterations required and achieves higher accuracy on the SCB-Dataset than existing loss functions. Our experimental results show that Wise-IoU v1, v2, and v3 all outperform existing loss functions, with Wise-IoU v3 achieving the best overall performance.

\section{Related Works}
\noindent\textbf{Student classroom behavior dataset}

Recently, many researchers have utilized computer vision to detect student classroom behaviors. However, the lack of public student behavior datasets in the education field restricts the research and application of behavior detection in classroom scenes. Many researchers have also proposed many unpublished datasets, such as Fu\cite{b13} et al, construct a class-room learning behavior dataset named as ActRec-Classroom, which include five categories of  listen, fatigue, hand-up, sideways and read-write with 5,126 images in total. And R Zheng\cite{b14} et al, build a large-scale student behavior dastaset from thirty schools, labeling these behaviors using bounding boxes frame-by-frame, which contains 70k hand-raising samples, 20k standing samples, and 3k sleeping samples. And Sun B\cite{b15} et al, presents a comprehensive dataset that can be employed for recognizing, detecting, and captioning students’ behaviors in a classroom. Author collected videos of 128 classes in different disciplines and in 11 classrooms.    However, the above datasets are from real monitoring data and cannot be made public.

\noindent\textbf{Students classroom behavior detection}

Mature object detection is used by more and more researchers in student behavior detection, such as YAN Xing-ya\cite{b4} et al. proposed a classroom behavior recognition method that leverages deep learning.   Specifically, they utilized the improved Yolov5 target detection algorithm to generate human detection proposals, and proposed the BetaPose lightweight pose recognition model, which is based on the Mobilenetv3 architecture, to enhance the accuracy of pose recognition in crowded scenes.   And ZHOU Ye\cite{b16} et al has proposed a method for detecting students' behaviors in class by utilizing the Faster R-CNN detection framework.   To overcome the challenges of detecting a wide range of object scales and the imbalance of data categories, the approach incorporates the feature pyramid and prime sample attention mechanisms.

\noindent\textbf{Attention mechanisms}
 
Attention is a crucial mechanism that can be utilized by various deep learning models in different domains and tasks.  The beginning of the attention mechanisms we use today is often traced back to their origin in natural language processing\cite{b17}.  The Transformer model proposed in \cite{b18} represents a significant milestone in attention research as it demonstrates that the attention mechanism alone can enable the construction of a state-of-the-art model.  Recently, sparse attention has gained popularity in the realm of vision transformers due to the remarkable success of the Swin transformer\cite{b19}.  Several works endeavor to make the sparse pattern adaptable to data, including DAT\cite{b20}, TCFormer\cite{b21}, and DPT\cite{Dpt}.  Additionally, BiFormer \cite{BiFormer} proposes a new dynamic sparse attention approach via bi-level routing to enable a more flexible allocation of computations with content awareness.

\noindent\textbf{Bounding Box Regression Loss}

Bounding box regression (BBR) plays an important role in object detection, particularly in the YOLO series, which has been widely acknowledged by many researchers and applied in various scenarios. The BBR loss in YOLOv1\cite{YOLOv1} and YOLOv2\cite{YOLOv2} is quite similar. In YOLOv2, the BBR loss is defined as $\left|\vec{B} - \vec{B_{gt}}\right|$, where $\vec{B}=[x\ y\ w\ h]$ represents the anchor box and the values within it correspond to the center coordinates and size of the bounding box. Similarly, $\vec{B_{gt}}=[x_{gt}\ y_{gt}\ w_{gt}\ h_{gt}]$ describes the properties of the target box. However, it cannot completely shield the interference of the size of the bounding box. YOLOv3\cite{YOLOv3} constructs $2-w_{gt}h_{gt}$ to alleviate this problem to some extent.

Intersection over Union \cite{IoU} (IoU) effectively balances the learning of large and small objects in BBR by shielding the interference of bounding box size through proportion. However, when there is no overlap between bounding boxes, the gradient back-propagated by IoU vanishes, preventing the width of the overlapping region from being updated during training. To solve this problem, several existing works (giou \cite{giou}, diou \cite{diou}, eiou \cite{eiou}, siou \cite{siou}) have constructed a penalty term that considers different geometric factors related to the bounding box. Especially, in Focal-EIoU v1 \cite{eiou}, Zhang et al. proposed using a non-monotonic FM to specify the boundary value of anchor boxes so that the box with the highest gain has an IoU equal to the boundary value. However, this approach does not fully exploit the potential of the non-monotonic FM or consider the quality evaluation of anchor boxes reflected in mutual comparison. In \cite{Wise-IoU}, Tong Z et al. proposed a dynamic FM $f(\beta)$ based on the estimated outlier degree of the anchor box. This approach enables BBR to focus on ordinary-quality anchor boxes by assigning small gradient gains to high-quality boxes with small $\beta$, while also reducing the impact of low-quality boxes with large $\beta$.

\section{SCB-Dataset}

Understanding student behavior is crucial for comprehending their learning process, personality, and psychological traits, and is important in evaluating the quality of education. The hand-raising, reading, writing behaviors are important indicators of evaluating classroom quality. However, the lack of public datasets poses a significant challenge for AI research in the field of education. To address this issue, we constructed a dataset that contains hand-raising, reading, and writing behaviors. This dataset presents unique characteristics and challenges due to the complexity and specificity of educational scenes, providing new opportunities for researchers. We collected nearly one thousand videos from kindergarten to high school, and selected 3 to 15 specific behavior video frames from each video, ensuring its representativeness and realism in reflecting the complexity of student behavior. The videos were extracted from the bjyhjy, 1s1k, youke.qlteacher and youke-smile.shec websites.

\begin{figure}
\centerline{\includegraphics[width=0.48\textwidth]{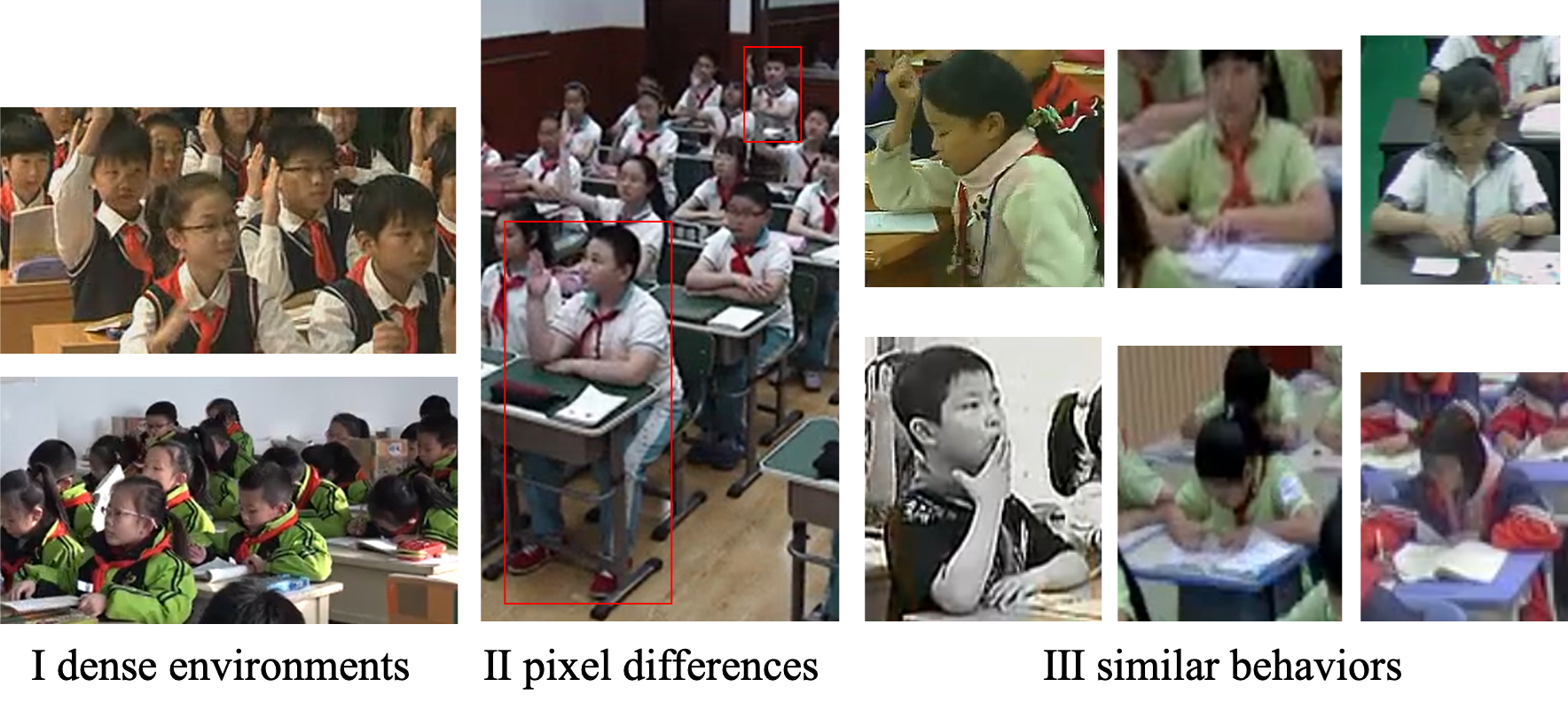}}
\caption{Challenges in the SCB-Dataset include dense environments, similar behaviors, and pixel differences..}
\label{dense_pixel_similar}
\end{figure}

The classroom environment presents challenges for detecting behavior due to the crowded students and variation in positions, as shown in Fig.~\ref{dense_pixel_similar} \uppercase\expandafter{\romannumeral1} and \uppercase\expandafter{\romannumeral2}. And  there is also visual similarity between hand-raising and other behavior classes, and in our experiments, we found that reading and writing also share a high degree of similarity in some scenes, further complicating detection as seen in Fig.~\ref{dense_pixel_similar} \uppercase\expandafter{\romannumeral3}.

\begin{figure}
\centerline{\includegraphics[width=0.48\textwidth]{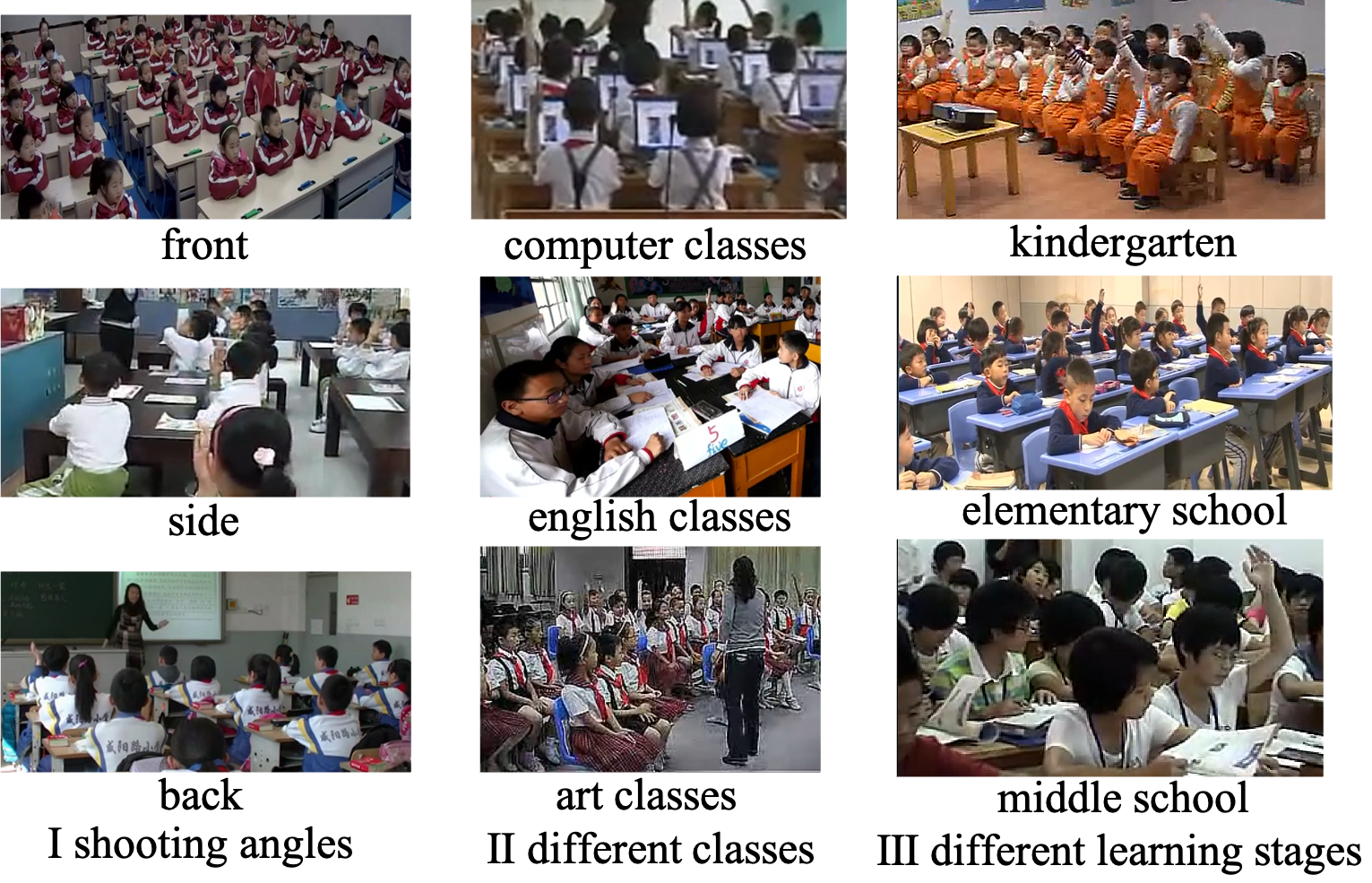}}
\caption{Challenges in the SCB-Dataset include varying shooting angles, class differences, and different learning stages.}
\label{angles_class_stages}
\end{figure}

The SCB-Dataset was collected from different angles, including front, side, and back views (Fig.~\ref{angles_class_stages} \uppercase\expandafter{\romannumeral1}). Additionally, the classroom environment and seating arrangement can vary, which adds complexity to the detection task (Fig.~\ref{angles_class_stages} \uppercase\expandafter{\romannumeral2}). Student classroom behavior behaviors also differ across learning stages from kindergarten to high-school, creating challenges for detection (Fig.~\ref{angles_class_stages} \uppercase\expandafter{\romannumeral3}).

We conducted a statistical analysis on the SCB-Dataset, which comprised 4.2k images with an average of 4.4 individuals with annotations in each image.

\section{YOLOv7+BRA}

Given the challenges of dense environments, pixel differences and similar student behaviors, we selected YOLOv7 as the foundational model due to its comprehensive consideration of speed and accuracy.   This lightweight one-stage object detection algorithm has an inference speed of 6.9 ms per image, allowing for a frame rate of over 140fps, which is suitable for real-time monitoring requirements in student classroom behavior detection.

Generally, one-stage object detection models can be divided into three parts: backbone, neck and head. The purpose of the backbone is to extract and select features, the neck is to fuse features, and the head is to predict results. However, YOLOv7 only retains the backbone and head parts because it proposes an Extended efficient layer aggregation networks (E-ELAN) module to replace various FPNs and PANs commonly used for feature fusion in the neck. Additionally, the Model scaling operation is common in concatenation-based models, which increases the input width of the subsequent transmission layer. Therefore, YOLOv7 proposes the compound scaling up depth and width method.

\begin{figure}
\centerline{\includegraphics[width=0.48\textwidth]{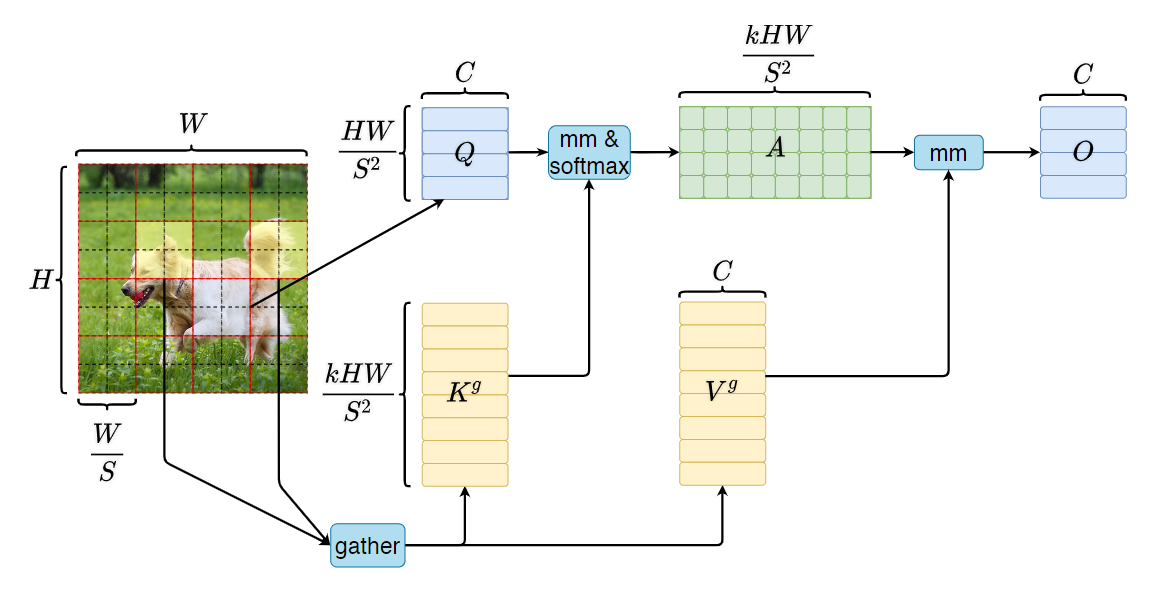}}
\caption{Bi-level Routing Attention.}
\label{BRA}
\end{figure}

Although YOLOv7 is considered one of the top object detection models, we found it challenging to handle occlusions and distinguish similar actions when detecting the SCB-Dataset.  To address these limitations, we introduced the bi-level routing attention (BRA) module, which is a novel dynamic sparse attention that enables more flexible computation allocation and content awareness (Fig.~\ref{BRA}).  This allows the model to have a dynamic query-aware sparsity.  BRA filters out irrelevant key-value pairs at a coarse region level, resulting in only a small portion of routed regions remaining.

Despite being one of the best object detection models available, we found that YOLOv7 struggled with handling occlusions and distinguishing similar actions when detecting the SCB-Dataset. Therefore, we introduced the bi-level routing attention (BRA) module to YOLOv7. BRA is a novel dynamic sparse attention that achieves more flexible computation allocation and content awareness, allowing the model to have dynamic query-aware sparsity. The key to BRA is filtering out most of the irrelevant key-value pairs at a coarse region level, so that only a small portion of routed regions remain. The whole algorithm is summarized with Torch-like pseudo code in Algorithm1

The process of BRA can be easily divided into three steps (Fig.~\ref{BRA}): firstly, assuming we input a feature map, we divide it into several regions, and obtain query, key, and value through linear mapping. Secondly, we use the adjacency matrix to build a directed graph to find the participating relationship corresponding to different key-value pairs, which can be understood as the regions that each given region should participate in. Finally, with the routing index matrix from region to region, we can apply fine-grained token-to-token attention.

The structure of our modified YOLOv7-BRA (YOLOv7 with Bi-level Routing Attention) model is shown in the Fig.~\ref{YOLOv7+BRA+wise}. We place the BRA module in the final part of the backbone. When introducing the BRA module, we considered placing it in three different positions: (1) replacing all convolutions with convolutions that include BRA; (2) placing BRA in the head section; and (3) placing BRA in the backbone section. If we choose method (1), it will result in a very large model that is difficult to train and affects inference speed. As for whether to place BRA in the head or in the backbone, considering that the role of the attention mechanism is to make the model only focus on specific areas of the image rather than the entire image, we believe this is part of feature extraction. Therefore, we chose to place it in the backbone section.

\section{YOLOv7+Wise-IoU}

The loss function for bounding box regression (BBR) is essential for object detection, and a well-defined BBR loss function can significantly improve the model's performance. Most existing works assume that the examples in the training data are of high-quality and focus on strengthening the fitting ability of the BBR loss.    However, blindly strengthening the fitting ability of BBR on low-quality examples, such as student classroom scenes where data quality can be uneven, will jeopardize localization performance. We adopt Wise-IoU to solve this problem. Wise-IoU is an IoU-based loss with a dynamic non-monotonic FM. The dynamic non-monotonic FM uses the outlier degree instead of IoU to evaluate the quality of anchor boxes and provides a wise gradient gain allocation strategy. This strategy reduces the competitiveness of high-quality anchor boxes while also reducing the harmful gradient generated by low-quality examples. This allows Wise-IoU to focus on ordinary-quality anchor boxes and improve the detector’s overall performance.

Wise-IoU v1 reduces geometric penalty when anchor box matches the target box well, improving model generalization with minimal training intervention. Wise-IoU v1 uses two layers of attention mechanisms, where the first layer is a distance attention function and the second layer amplifies the intersection over union of the ordinary-quality anchor box and reduces it for the high-quality anchor box. WIoU v1 is defined as:

\begin{equation}
    \wiou{1}~ = ~\r_{WIoU} \iou xc
    \label{wiou1_1}
\end{equation}

\begin{equation}
    \r_{WIoU}~ = ~\exp(\frac{\cenconnect}{(\diagsqr)^*})
    \label{wiou1_2}
\end{equation}

where $W_g$ and $H_g$ are the size of the smallest enclosing box, and $\diagsqr$ is the diagonal length of the smallest enclosing box. The superscript $^*$ indicates a detachment operation, which is used to prevent $\r_{WIoU}$ from producing gradients that hinder convergence.

Wise-IoU v2 introduces a monotonic focusing coefficient $\iou^{\gamma *}$ to Wise-IoU v1 that reduces the impact of simple examples on loss value to enable the model to focus on hard examples and enhance classification performance, and dynamically normalizes the focusing coefficient using the mean of $\iou$ over the batch. WIoU v2 is defined as:
\begin{equation}
    \wiou{2}~ = ~(\out)^{\gamma} \wiou{1}
    \label{wiou2}
\end{equation}

where $\out$ is the outlier degree of the anchor box, and $\gamma$ is a hyperparameter.

WIoU v3 uses a non-monotonic focusing coefficient based on the outlier degree $\beta$, which is defined as the ratio of $\iou$ to the mean of $\iou$ over the batch. WIoU v3 also dynamically updates the mean of $\iou$, and introduces a momentum $m$ to delay the time when the mean approaches the real value. WIoU v3 is defined as:

\begin{equation}
    \wiou{3}~ = ~r\wiou{1}
    \label{wiou3_1}
\end{equation}

\begin{equation}
    r~ = ~\frac{\beta}{\delta \alpha^{\beta - \delta}}
    \label{wiou3_2}
\end{equation}

where $C$ is a constant value, and $\delta$ makes $r=1$ when $\beta=\delta$.

We used Wise-IoU v1, v2 and v3 in yolov7 and tested their effect on the SCB-Dataset. 


\section{Experiment and Analysis}
\subsection{Experimental Environment and Dataset}
The experiment was compiled and tested using python 3.8, the corresponding development tool was PyCharm, the main computer vision library was python-OpenCV 4.1.2, the deep learning framework used was Pytorch v2.0.1 with CUDA version 11.7 for model training, the operating system was Ubuntu 20.04.2, and the CPU was a  20 CPU cores processor with Intel(R) Core(TM) i9-9900X.  The main frequency is 3.50GHz, the graphics card is  NVIDIA GeForce RTX 2080 Ti , the RAM is 11GB, and the hard disk capacity is 6TB SSD.

The dataset used in our experiments is SCB-Dataset, which we split into training, validation sets with a ratio of 4:1.

\begin{figure}
\centerline{\includegraphics[width=0.5\textwidth]{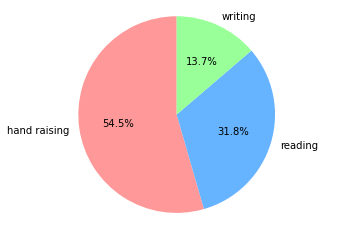}}
\caption{SCB-Dataset annotation distribution.}
\label{annotation_distribution}
\end{figure}

As shown in Fig.~\ref{annotation_distribution}, the SCB-Dataset  is affected from class imbalance. To solve this problem, YOLOv7 proposes a solution called class weight, with the following formula:

\begin{equation}
    w_c = \frac{1}{n_c}
    \label{w_c}
\end{equation}

Here, $w_c$ represents the weight of the $c$-th class, and $n_c$ represents the number of samples in the $c$-th class in the training dataset. The purpose of this formula is to balance the training difficulty of different classes and increase the importance of classes with fewer samples during training. This is achieved by assigning higher weights to classes with fewer samples during training, thereby improving their recognition performance.

\subsection{Analysis of experimental results}

When evaluating the results of an experiment, we use two main criteria: Precision and Recall. True Positive (TP) represents a correct identification, False Positive (FP) means an incorrect identification, and False Negative (FN) indicates that the target was missed.

To calculate Precision (Eq. \ref{precision}), we divide the number of True Positives by the sum of True Positives and False Positives. For Recall (Eq. \ref{recall}), we divide the number of True Positives by the sum of True Positives and False Negatives. Both Precision and Recall must be considered to properly assess the accuracy of the experiment.

\begin{equation}
    \centering
    \begin{aligned}
        precision~ = ~\frac{TP}{TP + FP}
    \end{aligned}
    \label{precision}
\end{equation}

\begin{equation}
    \centering
    \begin{aligned}
        recall~ = ~\frac{TP}{TP + FN}
    \end{aligned}
    \label{recall}
\end{equation}

 To provide a more thorough evaluation of Precision and Recall, the metrics of Average Precision (AP, Eq. \ref{AP}) and mean Average Precision (mAP, Eq. \ref{mAP}) have been introduced.  These metrics calculate the average Precision over a range of Recall values, which provides a more comprehensive assessment of the model's performance.  AP is the average of Precision values at all Recall levels, and mAP is the mean AP value averaged over different categories or classes.  
 
 \begin{equation}
    \centering
    \begin{aligned}
        AP_{i} = {\int_{0}^{1}{P(r)dr}}\\
    \end{aligned}
    \label{AP}
\end{equation} 

\begin{equation}
    \centering
    \begin{aligned}
        mAP = \frac{1}{n}{\sum\limits_{i}^{n}\left( AP_{i} \right)}
    \end{aligned}
    \label{mAP}
\end{equation}

As shown in Table \ref{tab:yolov7_bra_wiseiou_performance}, mAP50 (mAP@0.5) refers to the mean average precision at IoU threshold 0.5, while mAP5095 (mAP@0.5:0.95) represents the mean average precision at IoU thresholds ranging from 0.5 to 0.95. In addition, "W1," "W2," and "W3" respectively denote the Wise IoU versions 1, 2, and 3. We conducted a comprehensive comparison test to evaluate the performance of YOLOv7 with and without BRA, Wise Iou v1, Wise Iou v2, and Wise Iou v3. Our results indicate that our model outperforms the original YOLOv7. By combining our methods, the final mAP@0.5 has increased by 1.8$\%$, additionally, our methods have increased the mAP@0.5:0.95 and precision by 2.0$\%$ and 5.7$\%$, respectively, which demonstrates the effectiveness of our approach.

\begin{table}
    \centering \renewcommand\arraystretch{1.25} 
    \caption{comparisons of detection result between yolov7, YOLOv7+BRA and YOLOv7+Wise-IoU.}
    \label{tab:yolov7_bra_wiseiou_performance}
    \begin{tabular}{ccccc}
        \hline
        Method & precision & recall & mAP50 & mAP5095 \\ \hline
        YOLOv7 & 72.0$\%$ & \textbf{73.9$\%$} & 77.2$\%$ & 58.8$\%$ \\
        YOLOv7+BRA(ours) & 74.3$\%$ & 72.6$\%$  & 78.8$\%$ & 60.4$\%$ \\
        YOLOv7+W1(ours) & 75.0$\%$ & 73.2$\%$ & 78.7$\%$ & \textbf{60.8$\%$} \\
        YOLOv7+W2(ours) & 73.8$\%$  & 73.2$\%$  & 77.9$\%$ & 60.0$\%$  \\
        YOLOv7+W3(ours) & \textbf{77.7$\%$} & 70.4$\%$ & \textbf{79.0$\%$}  & 60.3$\%$ \\ 
        YOLOv7+BRA+W3(ours) & 75.0$\%$ & 73.8$\%$ & 78.9$\%$ & 60.2$\%$ \\ \hline
    \end{tabular}
\end{table}

\begin{figure}
\centerline{\includegraphics[width=0.45\textwidth]{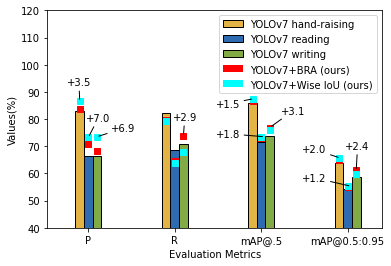}}
\caption{Comparison of YOLOv7, YOLOv7+BRA, and YOLOv7+Wise Iou in various evaluation metrics.}
\label{behavior_RPmAP}
\end{figure}

\begin{figure}
\centerline{\includegraphics[width=0.45\textwidth]{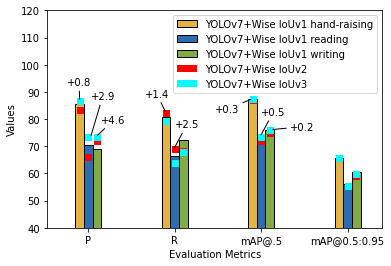}}
\caption{Comparison of YOLOv7+Wise Iou V1, YOLOv7+Wise Iou V2, and YOLOv7+Wise Iou V3 in various evaluation metrics.}
\label{behavior_RPmAP2}
\end{figure}

Regarding category-wise behaviors (Fig.~\ref{behavior_RPmAP}, Fig.~\ref{behavior_RPmAP2}), our model has shown improvement in most evaluation metrics compared to the original YOLOv7. The X-axis represents different evaluation metrics, with P representing precision and R representing recall, and each metric corresponds to the test results of the three types of behaviors in the figure.

In Fig.~\ref{behavior_RPmAP}, the bar chart represents the detection results of YOLOv7, which serves as the baseline for comparison with the detection results of YOLOv7+BRA and YOLOv7+Wise IoU (represented by square blocks), and the Wise IoU is based on the v3 version, it is evident that, except for the recall metric, both YOLOv7+BRA and YOLOv7+Wise IoU exhibit significant improvements over YOLOv7. Particularly, the precision metric shows the most noticeable improvement, with hand-raising, reading, and writing behaviors exhibiting a respective increase of 3.5$\%$, 7.0$\%$, and 6.9$\%$.

In Fig.~\ref{behavior_RPmAP2}, we compared the detection performance of Wise IoU versions 1, 2, and 3. The bar chart represents the detection results of YOLOv7+Wise IoU v1, which serves as the baseline for comparison with the detection results of YOLOv7+Wise IoU v2 and YOLOv7+Wise IoU v3 (represented by square blocks). From the figure, it is evident that the v2 version generally outperforms the v1 version, and the v3 version generally outperforms the v2 version.

\begin{figure}
\centerline{\includegraphics[width=0.45\textwidth]{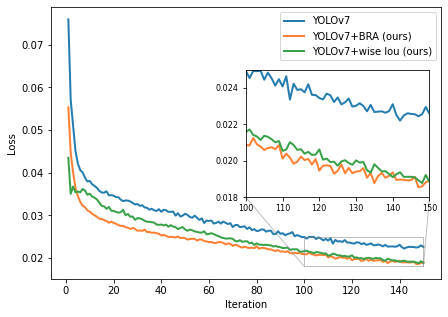}}
\caption{Comparison of YOLOv7, YOLOv7+BRA, and YOLOv7+Wise Iou in training set Bounding Box loss.}
\label{BoundingBoxLoss}
\end{figure}

From Fig.~\ref{BoundingBoxLoss}, it can be observed that the loss curves of YOLOv7+BRA and YOLOv7+Wise IoU converge faster, which improves the training speed of the models while achieving lower loss rates.

\begin{figure}
\centerline{\includegraphics[width=0.45\textwidth]{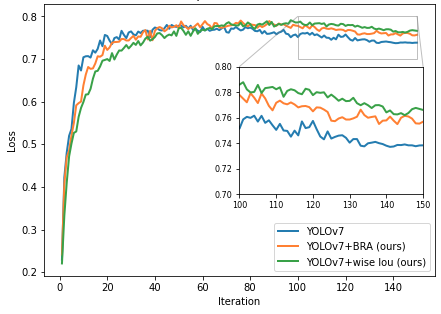}}
\caption{Comparison of YOLOv7, YOLOv7+BRA, and YOLOv7+Wise Iou in mAP@.5.}
\label{mAP05}
\end{figure}

From Fig.~\ref{mAP05}, it is evident that YOLOv7+BRA and YOLOv7+Wise IoU achieve higher mAP@0.5, indicating that the modified student behavior recognition model is more accurate and suitable for use in educational settings.

\begin{figure*}
\centerline{\includegraphics[width=0.69\textwidth]{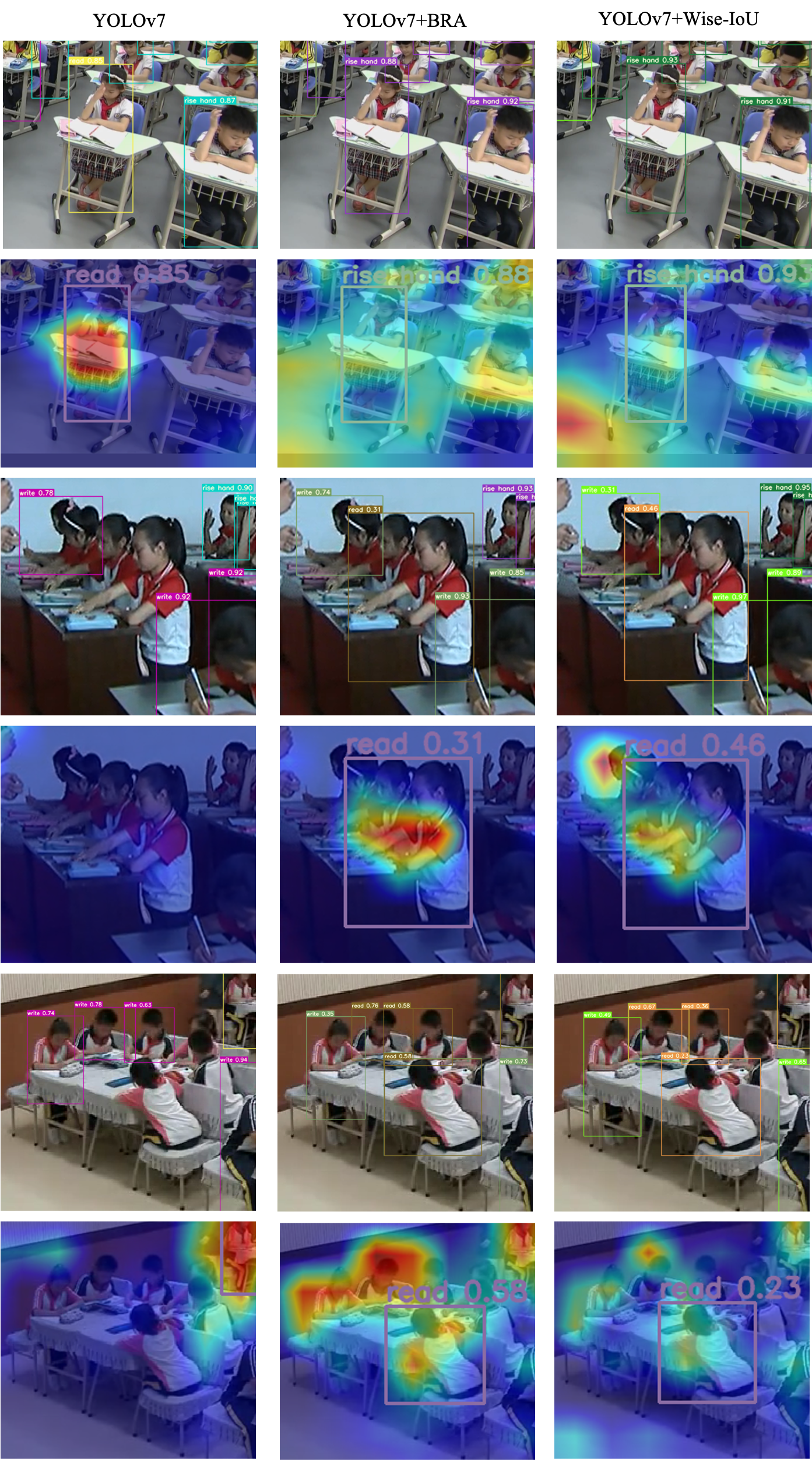}}
\caption{SCB-Dataset annotation distribution.}
\label{gradcam}
\end{figure*}

To observe the neural network's contribution to the final prediction, we used Grad-CAM\cite{Grad-cam} to visualize the decision process of the second-to-last layer of the network, as shown in Fig.~\ref{gradcam}. It can be seen that YOLOv7+BRA and YOLOv7+Wise IoU show improved detection performance in crowded scenes and detecting small objects far back in the classroom, compared to YOLOv7. Furthermore, the visualized network shows that YOLOv7+BRA and YOLOv7+Wise IoU can better direct the network's attention to the book area of the student (during reading and writing behavior). Through our extensive observations, we found that YOLOv7+Wise IoU outperforms YOLOv7+BRA in terms of precision, mAP@0.5, and mAP@0.5:0.95. However, the detection results of YOLOv7+Wise IoU exhibit more overlapping detections, as seen in row 5, column 3 of Fig.~\ref{gradcam}. In contrast, although the accuracy of YOLOv7+BRA+Wise IoU is slightly lower than that of YOLOv7+Wise IoU, it has effectively minimized overlapping detections in our experiments.

\section{Conclusion}

In summary, this paper proposes a Student Classroom Behavior Detection method based on an improved YOLOv7 for accurately detecting student behavior in classroom videos. We constructed the SCB-Dataset, including three behaviors, which fills a gap in current research on detecting student behavior in classroom teaching scenes. We integrated the biformer attention module and Wise-IoU into the YOLOv7 network to improve detection accuracy, achieving an mAP@0.5 of 79$\%$, resulting in a 1.8$\%$ improvement over previous results. Our experimental results show that our model outperforms the original YOLOv7 in terms of precision, mAP@0.5, and mAP@0.5:0.95. The proposed Student Classroom Behavior Detection method has important applications in education and can aid in improving teaching effectiveness. Overall, the results demonstrate the effectiveness of our approach and provide a new solution for behavior detection in educational settings.

\vspace{12pt}

\end{document}